\documentclass[runningheads]{llncs}

\usepackage{eccv}

\usepackage{eccvabbrv}

\usepackage{graphicx}
\usepackage{booktabs}

\usepackage[accsupp]{axessibility}  

\def\eg{\emph{e.g.}} 
\def\ie{\emph{i.e.}}

\def\etc{\emph{etc.}}

\def\etal{\emph{et al.}}
\def\vs{\emph{vs.}}

\usepackage{multirow}
\usepackage{bbding}
\usepackage{makecell}
\usepackage{colortbl}
\definecolor{mygray}{gray}{.9}
\usepackage{adjustbox}
\usepackage{hyperref}

\usepackage{hyperref}

\usepackage{orcidlink}

\begin{document}

\title{{\color{red}E}{\color{blue}v}Sign: Sign Language Recognition and Translation with Streaming Events} 

\titlerunning{{\color{red}E}{\color{blue}v}Sign: Sign Language Recognition and Translation with Streaming Events}

\author{Pengyu Zhang\inst{1,2\footnotemark[1]}
\and
Hao Yin\inst{1} 
\and
Zeren Wang\inst{1}
\and
Wenyue Chen\inst{1} \and
\\
Shengming Li\inst{1} 
\and
Dong Wang\inst{1}
\and
Huchuan Lu\inst{1}
\and
Xu Jia\inst{1\footnotemark[4]} 
}

\renewcommand{\thefootnote}{\fnsymbol{footnote}}
\footnotetext[1]{Work was done at Dalian University of Technology.}
\footnotetext[4]{Corresponding author: jiayushenyang@gmail.com}

\authorrunning{P. Zhang et al.}

\institute{Dalian University of Technology \and
National University of Singapore}
\maketitle

\renewcommand\thefootnote{\arabic{footnote}}

\begin{abstract}
Sign language is one of the most effective communication tools for people with hearing difficulties.
Most existing works focus on improving the performance of sign language tasks on RGB videos, which may suffer from  degraded recording conditions, such as fast movement of hands with motion blur and textured signer's appearance.
The bio-inspired event camera, which asynchronously captures brightness change with high speed, could naturally perceive dynamic hand movements, providing rich manual clues for sign language tasks.
In this work, we aim at exploring the potential of event camera in continuous sign language recognition~(CSLR) and sign language translation~(SLT). To promote the research, we first collect an event-based benchmark \textbf{EvSign} for those tasks with both gloss and spoken language annotations. EvSign dataset offers a substantial amount of high-quality event streams and an extensive vocabulary of glosses and words, thereby facilitating the development of sign language tasks.
In addition, we propose an efficient transformer-based framework for event-based SLR and SLT tasks, which fully leverages the advantages of streaming events. The sparse backbone is employed to extract visual features from sparse events. Then, the temporal coherence is effectively utilized through the proposed local token fusion and gloss-aware temporal aggregation modules.
Extensive experimental results are reported on both simulated~(PHOENIX14T) and EvSign datasets. Our method performs favorably against existing state-of-the-art approaches with only {\bf 0.34\%} computational cost~(0.84G FLOPS per video) and {\bf 44.2\%} network parameters. The project is available at \href{https://zhang-pengyu.github.io/EVSign/}{https://zhang-pengyu.github.io/EVSign}.

\end{abstract}

\section{Introduction}

\begin{figure}[t]
    \centering
    \includegraphics[width=1.0\textwidth]{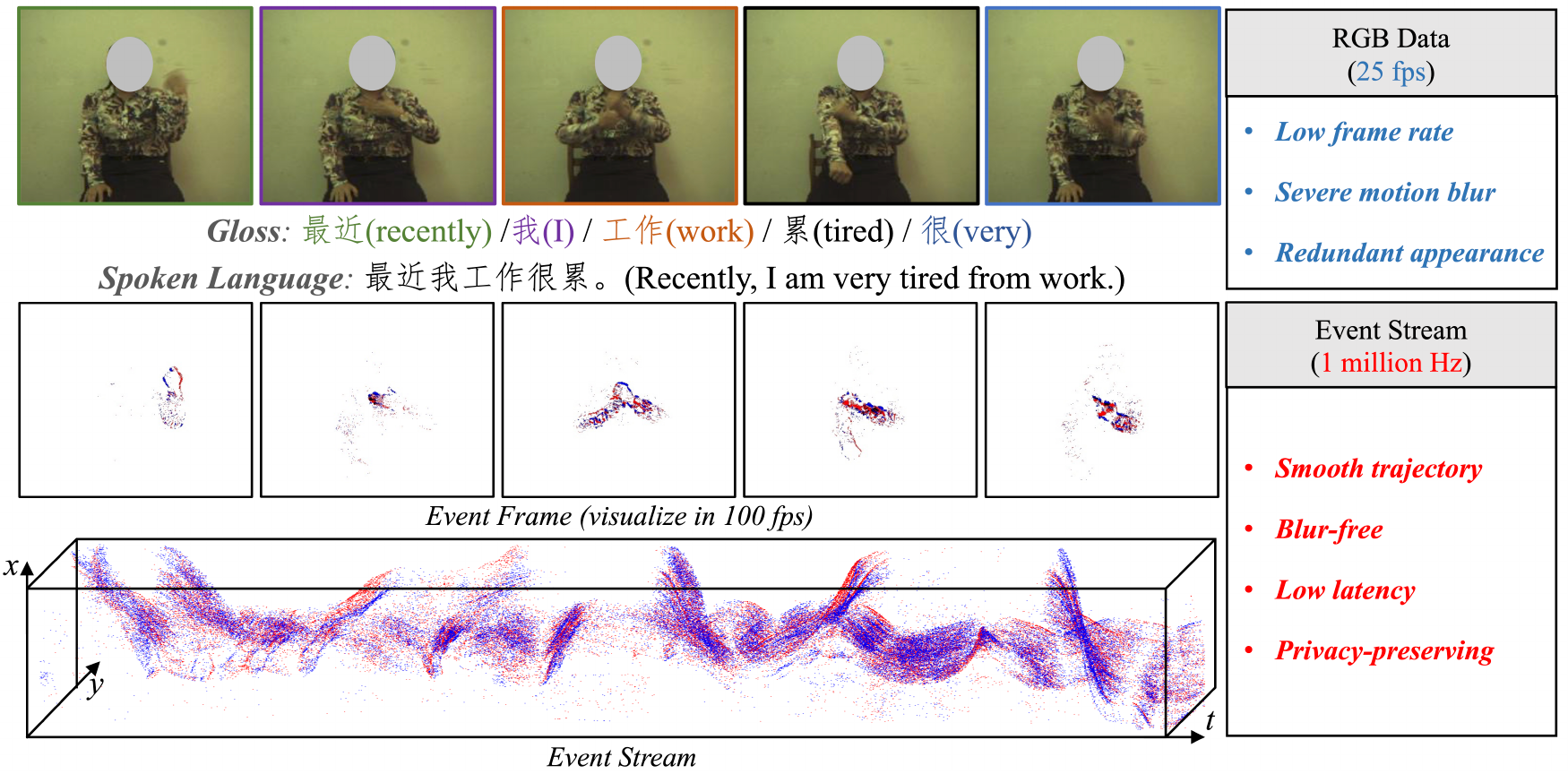}
    \caption{Comparison between sign language recognition and translation with RGB and event data. We provide the first benchmark for event-based CSLR and SLT tasks, namely EvSign. Compared with RGB data, event stream can capture smooth movement within microsecond-level response, avoiding motion blur. Furthermore, the sparse event only stresses on the moving targets, such as hands and arms, which can be processed efficiently and protects personal privacy~(facial information).}
    \label{fig1}
    \vspace{-3mm}
\end{figure}

As a main communication medium employed by the deaf community, sign language conveys multi-cue information by manual features~(\eg, hand shape, movement and pose) and non-manuals features, including facial expression, mouth gesture and other body movements~\cite{Ong_PAMI05_HandCraft2,Penny_book01_SLsurvey2}.
According to the outputs, sign language based tasks can be mainly categorized into sign language recognition~(SLR) and sign language translation~(SLT). In SLR, the minimal lexical component, namely gloss, is predicted. SLR can be further categorized into isolated SLR~(ISLR) and continuous SLR~(CSLR). 
The purpose of SLT is to fully translate sign language into spoken language, which is often considered as a sequence-to-sequence learning problem. Recent works are based on videos captured by conventional frame-based sensors, which suffer from challenging scenarios, including severe motion blur and cluttering distractors. As shown in Fig.~\ref{fig1}(a), the information will be degraded in extreme conditions, such as fast hand and arm movement, thereby leading to limited performances.
To this end, existing works~\cite{Guo_CVPR23_DCTCCSLR,Min_ICCV21_VAC,Cui_TMM19_IT, Zhou_AAAI20_STMC,Hu_CVPR23_CorrNet} emphasize temporal modeling, which leverages temporal cues in pixel- and frame-wise representations via 3D convolution and Long-Short Term Memory~(LSTM) networks. 
Furthermore, STMC~\cite{Zhou_AAAI20_STMC} and CorrNet~\cite{Hu_CVPR23_CorrNet} are designed to construct discriminative spatial representation,  focusing on body trajectories.
%

Event camera, a biologically-inspired sensor, detects the variation of intensity along time. Rather than encoding visual appearance with still images, it generates sparse and asynchronous event stream with extremely high temporal resolution~\cite{Brandli_JSSC14_1MHz}~(1M Hz \vs 120 Hz), high dynamic range and low latency, which is ideally suited for extracting motion cues~\cite{Anton_IROS18_evmotion1,Zhu_RSS18_EvFlowNet,Zhang_ICCV21_FE108}. 
As shown in Fig.~\ref{fig1}(b), event camera could benefit sign language tasks from four perspectives. First, event data can capture richer motion information, thereby facilitating limb movements modeling effectively. Second, conventional images may be degraded due to the rapid movements, while event camera can record the sharp boundary for further processing.
Third, the event stream contains less redundant information, such as background and textured clothing, which can boost efficiency and avoid distractor interference. Fourth, from a privacy protection perspective, event cameras can avoid collecting static facial information.
There have been several attempts~\cite{Vasudevan_PAA22_SLAnimals1,Vasudevan_FG20_SLAnimals,Wang_WASA21_EvASL,Shi_SPL23_QISampling} to leverage event camera for sign language tasks. However, current event-based sign language datasets~\cite{Vasudevan_FG20_SLAnimals,Vasudevan_PAA22_SLAnimals1,Wang_WASA21_EvASL} only provide sign videos for ISLR. The limited vocabulary size and frame length cannot meet the requirements of real-world applications. Furthermore, the designed methods~\cite{Shi_SPL23_QISampling,Wang_WASA21_EvASL} are based on the networks originally designed for frame sequences, such as AlexNet~\cite{Alex_NIPS12_AlexNet} and ResNet~\cite{He_CVPR16_ResNet}, which do not fully leverage the advantage of event data.


To unveil the power of event-based sign language tasks, we collect an event-based Chinese sign language benchmark, namely EvSign. To the best of our knowledge, it is the first dataset designed for event-based CSLR and SLT tasks. More than 6.7K high-resolution event streams are collected in EvSign, which is of comparable scale to the existing RGB-based SLR datasets. The large corpus with native expressions, precise gloss and spoken language annotations can promote the development of CSLR and SLT tasks. 

Moreover, we present a transformer-based framework to make full use of the advantage of event data. 
First, a sparse backbone is employed to efficiently compute visual features on event data to obtain visual features efficiently. Then, temporal information is modeled via the proposed local token fusion and gloss-aware temporal aggregation modules, where the visual tokens are firstly combined to model local motion and reduce the computational cost. 
Subsequently, we aggregate the temporal information during the whole video into fused tokens hierarchically, which is then used for gloss and word prediction. 
Our method achieves very competitive performance on both synthetic and real datasets for both tasks with only 0.34\% computational cost.

To sum up, our contributions can be concluded as three aspects:
\begin{itemize}
    \item We propose the first benchmark for event-based CSLR and SLT tasks, which contains high-quality event streams, comprehensive corpus and precise gloss and spoken language annotation.
    \item We design a transformer-based algorithm for both tasks, which fully leverages the characteristics of event data. 
    \item Experiments on both synthesized PHOENIX14T and EvSign datasets demonstrate that our method achieves favorable performance with only 0.34\% FLOPS and 44.2\% parameters against existing algorithms.
\end{itemize}

\section{Related work}

\subsection{RGB-based sign language recognition}
Sign language recognition can be categorized into two main directions: isolated SLR~(ISLR)\cite{Hu_ICCV21_SignBERT,Hu_AAAI21_HMM-CNN} and continuous SLR~(CSLR)~\cite{Camgoz_CVPR20_SLT,Zuo_CVPR22_C2SLR,Zhou_AAAI20_STMC,Hao_ICCV21_SMDL}. 
As for ISLR, word-level prediction is performed, while CSLR aims to predict a series of glosses from longer-term videos, which has become the primary focus of research due to its closer alignment with real-world applications.
To tackle CSLR, researchers mainly work on extracting discriminative spatial and temporal information.
In the early stage, hand-crafted features~\cite{Freeman_FG95_HandCraft1,Ong_PAMI05_HandCraft2} are utilized to extract spatio-temporal representation. HMM-based algorithms~\cite{Koller_CVPR16_HMM1,Koller_CVPR17_HMM2,Koller_BMVC16_HMM3,Koller_PAMI20_WSLSLV} are designed to predict gloss progressively.
Recently, deep-based frameworks have been proposed, focusing on leveraging motion-aware information~\cite{Hu_AAAI21_HMM-CNN,Pu_CVPR19_IANSLR,Zhou_AAAI20_STMC,Niu_ECCV20_SFLMSGC}. 
CTC loss~\cite{Graves_ICML06_CTC} is adopted to temporally align the classification results with unsegmented sequences, achieving end-to-end training.
In addition, some works explore the use of other modalities to provide a complementary cue to visual signals, including skeleton~\cite{Chen_NIPS22_TwoStream, Zuo_CVPR22_C2SLR, Parelli_ICASSP22_skeleton2,Jiao_ICCV23_skeleton3,Duarte_CVPR21_How2Sign} and depth~\cite{Oszust_KES21_depth1,Zheng_icarcv16_depth2,Kumar_NeuroComputing17_depth3}, \etc~In this paper, we exploit the effectiveness of a bio-inspired sensor, \ie, event camera, for CSLR task.



\subsection{RGB-based sign language translation}
Compared to SLR, Sign Language Translation~(SLT) aims to generate spoken language translations from sign language videos in a progressive manner.
Camgoz \etal~\cite{Necati_CVPR18_PHOENIX14T} first introduce SLT task and formalize it into Neural Machine Translation in an end-to-end setting, which extracts gloss features through a CNN model and SLT using a sequence-to-sequence model. Subsequent works~\cite{Camgoz_CVPR20_SLT,Zhou_CVPR21_CSLDaily,Chen_CVPR22_SMTL,Li_NeurIPS20_TSPNet,Zhang_ICLR23_SLTUNET} have focused primarily on how to better extract spatial and temporal features. Some studies have attempted gloss-free methods~\cite{Yin_CVPR23_GlossFree,Lin_ACL23_GlossFreeSLT, Wong_ICLR24_Sign2GPT} to generate sentences without relying on gloss-level features, extensive experiments have shown that directly implementing an end-to-end Sign2Text model yields inferior results compared to using glosses as the intermediate supervision in the Sign2Gloss2Text model. Therefore, the current implementation of SLT task is predominantly based on the Sign2Gloss2Text approach.  Recent works mainly focus on vision based SLT, while SLT with other modalities have not been fully exploited. 

\subsection{Event-based sign language recognition}
Event camera captures the intensity variation of each pixel, recording the trajectory of fast-moving objects at high temporal resolution. Due to its  
property, it can provide sufficient temporal information, which is suitable for modeling object motion. A few attempts contribute to  ISLR~\cite{Shi_SPL23_QISampling, Wang_WASA21_EvASL,Vasudevan_FG20_SLAnimals,Vasudevan_PAA22_SLAnimals1}. 
Vasudevan \etal~\cite{Vasudevan_FG20_SLAnimals,Vasudevan_PAA22_SLAnimals1} propose an event-based Spanish sign language dataset, namely SL-Animals-DVS, consisting of 1,102 event streams regarding animals. Two Spiking Neural Networks~(SNN)~\cite{Shrestha_NIPS18_SLAYER,Wu_Arxiv17_STBP} are used for evaluation. Wang \etal~\cite{Wang_WASA21_EvASL} consider the event camera as a novel sensor in ISLR and collect an American sign language dataset, which contains 56 words. Shi \etal~\cite{Shi_SPL23_QISampling} design an event sampling strategy to select key event segments according to event distribution. The selected events are then fed to a CNN to obtain classification results. They also provide a synthetic dataset N-WLASL, where the event is collected by shooting an LCD monitor to record the videos from WLASL~\cite{Li_WACV20_WLASL}. 
Above all, three main challenges limit the development of SLR. 
First, existing benchmarks are in small vocabulary, which cannot fully exploit the potential of sign language recognition. Second, the event data is collected using the out-of-date sensors with low spatial resolution in those datasets, leading to missing details in hand gesture and subtle movement. Third, all the datasets are designed for ISLR. It can solely be used for specific applications and cannot be generalized in real scenes.
Therefore, it is crucial to collect a larger-scale dataset to promote the development of sign language tasks.

\section{EvSign benchmark}

\begin{table*}[t]
\caption{Summary of existing sign language recognition and translation benchmarks.}\label{dataset_comparison}
\vspace{-3mm}
\scriptsize
\begin{center}
	\begin{adjustbox}{width=1.0\textwidth,center}
    \begin{tabular}{lcccccccccc}
    \toprule
    Dataset & Lang. & \makecell{Gloss \\ Vocab.} & \makecell{Text \\ Vocab.} & \makecell{Num. \\ Videos} & \makecell{Num. \\ Signer}  & Continuous & SLT & Resolution & Source\\
    \midrule 
    WLASL~\cite{Joze_BMVC19_MSASL} & ASL & 1,000 & -- & 25,513 & 222 & {\tiny \XSolid} & {\tiny \XSolid} & Variable & Web\\
    DEVISIGN~\cite{Wang_TAC16_DEVISIGN} & CSL & 2,000 & -- & 24,000 & 8 & {\tiny \XSolid} & {\tiny \XSolid} &  598 $\times$ 448 & Lab\\ 
    PHOENIX-14~\cite{Koller_CVIU15_PHOENIX14} & DGS & 1,081 & -- & 6,841 & 9 & \checkmark & {\tiny \XSolid} & 210 $\times$ 260 & TV \\
    CCSL~\cite{Huang_AAAI18_CCSL} & CSL & 178 & -- & 25,000 & 50 & \checkmark & {\tiny \XSolid} & 1280 $\times$ 720 & Lab\\
    SIGNUM~\cite{Agris_07_SIGNUM} & DGS & 455 & -- & 15,075 & 25 & \checkmark & {\tiny \XSolid} & 776 $\times$ 578 & Lab \\ 
    PHOENIX-14T~\cite{Necati_CVPR18_PHOENIX14T} & DGS & 1,066 & 2,887 & 8,257 & 9 & \checkmark & \checkmark & 210 $\times$ 260 & TV \\
    CSL-Daily~\cite{Zhou_CVPR21_CSLDaily} & CSL & 2,000 & 2,343 & 20,654 & 10 & $\checkmark$ & $\checkmark$ & 1920 $\times$ 1080 & Lab \\
    Youtube-ASL~\cite{Uthus_arxiv23_youtubeASL} & ASL & -- & 60,000 & 11,093 & 2,519 & {\tiny \XSolid} & \checkmark &  Variable & Web\\
    \midrule
    EvASL~\cite{Wang_WASA21_EvASL} & ASL & 56 & -- & 11,200 & 10 & {\tiny \XSolid} & {\tiny \XSolid} & 128 $\times$ 128 & Lab\\
    SL-Animals-DVS~\cite{Vasudevan_FG20_SLAnimals} & SSL & 19 & -- & 1,121 & 59 & {\tiny \XSolid} & {\tiny \XSolid} & 128 $\times$ 128 & Lab\\
    \rowcolor{mygray} \bf EvSign(Ours) & CSL & 1,387 & 1,947 & 6,773 & 9 & $\checkmark$ & $\checkmark$ & 640 $\times$ 480 & Lab \\
    \bottomrule
    \end{tabular}
\end{adjustbox}
\end{center}
\vspace{-4mm}
\end{table*}

\subsection{Benchmark Statistics}
We use the DVXplorer-S-Duo camera from iniVation, which is a binocular camera capable of simultaneously capturing both event and RGB data. The spatial size of event stream is 640 $\times$ 480. We also record the RGB data with the size of 480 $\times$ 320 at 25 FPS for visualization and annotation.

To fit the practical usage, the corpus is sourced around daily life, such as shopping, education, medical care, travel and social communication, etc. The glosses are sampled from the Chinese national sign language dictionary~\cite{CNSLD} and CSL-Daily~\cite{Zhou_CVPR21_CSLDaily}, and are then reorganized into a spoken sentence. To avoid differences in expression, we further provide glosses and sentences to signers for adjustment to suit the deaf community.

We recruit 9 professional volunteers from the deaf community, who are familiar with general sign language for data collection.
We employ a two-step manner to avoid data ambiguity. When collecting sign data, the signers first watch a reference video and then start to perform the action. After recording, other three signers vote to determine whether the sign expression is precise and easy to understand. For each sample in the corpus, there are about three signers to perform the action.
We separate the sign videos into training, development and test subsets, which contain 5,570, 553 and 650 clips, respectively. As shown in Table~\ref{dataset_comparison}, the proposed dataset significantly surpasses existing datasets in vocabulary size, task scope, and data resolution, which provides a comprehensive corpus to exploit the power of event data in handling sign language tasks.

\subsection{Annotation}
In EvSign, both sign gloss and spoken language annotations are provided.
First, annotators identify all the glosses according to~\cite{CNSLD} in the RGB videos. We note that several signs may express the same meaning. Thus, the authors further revise the annotation to ensure that each sign language corresponds to a unique gloss annotation. Finally, the spoken language annotations are updated according to the gloss annotation. We employ tokenization method in HanLP~\footnote[1]{https://github.com/hankcs/HanLP} to separate a sentence into words.
As shown in Table~\ref{annotation_description}, EvSign provides 1.3K unique signs and 1.8K words, which cover various aspects of our daily life. Furthermore, more than 35K and 53K gloss and text annotations are totally labeled. 
\begin{table}[t]
\caption{Annotation statistics of EvSign dataset.}
\label{annotation_description}
\vspace{-3mm}
\scriptsize
\begin{center}
	\begin{adjustbox}{width=0.95\textwidth,center}
    \begin{tabular}{cccccccccc}
    \toprule
    & & Segments & Frame & Duration(h) & Vocab. & \makecell{Avg.\\words} & \makecell{Tot.\\words} & OOVs & Singletons\\
    \midrule
    \multirow{3}{*}{\makecell{Sign \\ Gloss}} & Train & 5,570 & 606.7K & 6.74 & 1,348 & 5.09 & 28,387 & -- & 230 \\
    & Dev & 553 & 75.1K & 0.83 & 695 & 5.68 & 3,145 & 25 & --\\
    & Test & 650 & 87.2K & 0.97 & 723 & 5.48 & 3,562 & 18 & --\\
    \midrule
    \multirow{3}{*}{Chinese} & Train & \multirow{3}{*}{$\uparrow$ same} & \multirow{3}{*}{$\uparrow$ same} & \multirow{3}{*}{$\uparrow$ same} & 1,825 & 7.76 & 43,276 & -- & 407\\
    & Dev & & & & 880 & 8.58 & 4,746 & 79 & --\\
    & Test & & & & 912 & 8.23 & 5,350 & 53 & --\\
    \bottomrule
    \end{tabular}
\end{adjustbox}
\end{center}
\vspace{-4mm}
\end{table}
\subsection{Evaluation Metrics}
We provide two evaluation protocols for both SLR and SLT. As for SLR evaluation, we use Word Error Rate~(WER) as the metric, which is widely used in sign language and speech recognition. WER measures the similarity of reference and hypothesis, which is based on the minimum number of operations required to convert the prediction into the reference sentence as:
\begin{equation}
    \text{WER} = \frac{\#\text{sub} + \#\text{ins} + \#\text{del}}{\#\text{ref}}
\end{equation}
where $\#\text{sub}$, $\#\text{ins}$ and $\#\text{del}$ are the number of basic operations, including substitution, insertion and deletion. $\#\text{ref}$ represents the number of words in the reference sentence. Lower WER indicates better performance. For SLT evaluation, we employ ROUGE~\cite{Lin_ACL04_ROUGE} and BLEU~\cite{Papineni_ACL02_BLEU} as evaluation metrics. Here, BLEU is calculated with n-grams from 1 to 4 and ROUGE-L~\cite{Lin_ACL04_ROUGEL} is used as our metric. The higher the ROUGE and BLEU scores, the better the performance.

\section{Methodology}

\begin{figure}[t]
    \centering
    \includegraphics[width=1.0\textwidth]{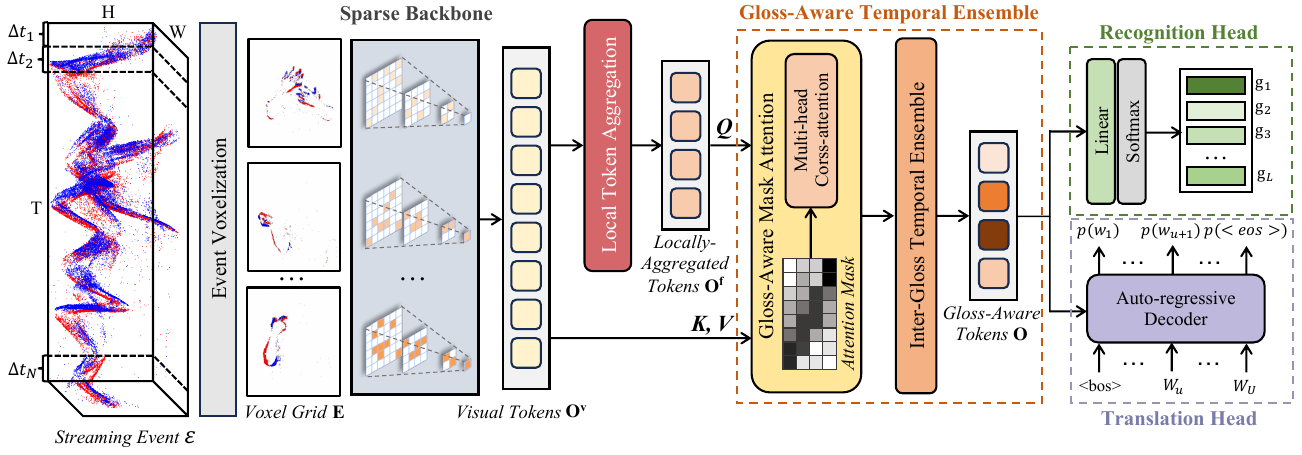}
    \caption{Pipeline of the transformer-based framework for CSLR and SLT tasks.} 
    \label{fig2}
\end{figure}

\subsection{Overview}
As shown in Fig.~\ref{fig2}, the event stream ${\cal E} = \{ e_i \}_{i=1}^N$ is firstly split to $P$ segments evenly and converted to a set of event representation $\mathbf{E} = \{\mathbf{E}_1, \mathbf{E}_2,..., \mathbf{E}_P \} \in \mathbb{R}^{P \times B \times H \times W}$, \ie, voxel grid~\cite{Zhu_CVPR19_VG3}, where $B$ denotes the bin size in voxel grid. Taken $\bf E$ as input, the proposed network is to jointly predict all the glosses $\mathcal{G} = \{g_1, g_2,...,g_Z\}$ and translate them into spoken language $\mathcal{W} = \{w_1, w_2,...,w_U\} $ in a sequence-to-sequence manner. $Z$ and $U$ are the number of glosses and words in the sign language video.
Our method contains five main parts, including sparse backbone~(SConv), Local Token Fusion~(LTF), Gloss-Aware Temporal Aggregation~(GATA) and two task heads, \ie, recognition head and translation head. First, SConv generates a set of visual tokens. Then, we employ LTF to fuse local motion from adjacent timestamps thereby reducing the number of tokens. Moreover, the temporal information is decoupled into intra-gloss and inter-gloss cues, which are learned hierarchically by GATA module. Specifically, we propose a gloss-aware mask attention to dynamically fuse the comprehensive motion information from visual tokens into the fused tokens. It measures the token's similarity in time and feature spaces, which can be aware of various action lengths. Furthermore, the global coherence among tokens from different glosses is learned via inter-gloss temporal aggregation, thereby obtaining the gloss-aware tokens. Finally, those tokens are sent to recognition and translation heads to predict the probability of target gloss sequence and spoken language.

\subsection{Overall framework}

\noindent \textbf{Sparse backbone~(SConv).} Due to the sparse property of event data, we build a sparse convolutional network~\cite{Liu_CVPR15_Spconv,Graham_Arxiv17_submSpconv} with the architecture of ResNet18~\cite{He_CVPR16_ResNet}. The backbone can process the event representation to obtain the visual tokens ${\bf O}^{\text v} = \{ {\bf o}^{\text v}_1, {\bf o}^{\text v}_2,..., {\bf o}^{\text v}_P \} \in \mathbb{R}^{P \times C}$. The sparse backbone is able to fully leverage the characteristics of data sparsity, thus significantly reducing the computational load. Compared to regular convolution layers, sparse backbone can also better maintain the sparsity at feature level, leading to a sharper boundary~\cite{Graham_Arxiv17_submSpconv}.

\noindent \textbf{Local Token Fusion~(LTF).}
Local motion integration is crucial for long-term temporal modeling, which can
first build an effective representation for continuous actions within a short duration and reduce computational load~\cite{Hu_ECCV22_TLP}. To construct a powerful features for local motion, we introduce LTF to fuse neighboring visual tokens, which contains two multi-head self-attention within local window~(W-MSA)~\cite{Liu_ICCV21_SwinTransformer} and two max pooling~(MaxPool) layers. A regular window partitioning scheme is adopted to all the visual tokens, where each window covers $I$ tokens. The self-attention within non-overlapping window is calculated to aggregate local motion information. Then, we introduce a max pooling operation to reduce the number of tokens. We adopt another W-MSA and MaxPool with the same window size and ratio to capture a longer-term movement, thus obtaining the fused tokens ${\bf O}^\text{f} \in \mathbb{R}^{L \times C}$. The LTF can be formulated as,
\begin{equation}
    \begin{aligned}
    &{\widetilde{\bf O}}^{\text v} = \text{MaxPool}(\text{W-MSA}({\bf O}^{\text v}) + {\bf O}^{\text v})\\
    &{\bf O}^\text{f} = \text{MaxPool}(\text{W-MSA}({\widetilde{\bf O}}^{\text v}) + {\widetilde{\bf O}}^{\text v})
    \end{aligned}
\end{equation}
where the number of tokens $L = \frac{P}{\gamma}$, while $\gamma$ represents the downsampling ratio.

\noindent\textbf{Gloss-Aware Temporal Aggregation~(GATA).} Global temporal modeling is a key step to exploit the correspondence of continuous signs in long-term videos.
Existing methods~\cite{Hu_AAAI21_HMM-CNN,Min_ICCV21_VAC,Hu_CVPR23_CorrNet} learn the global motion cue by 1D temporal convolution and BiLSTM, which ignores the varying durations of different signs. Simply applying temporal modeling among the fixed frames will learn a non-optimal representation and involve redundant information from different glosses.  
In this work, we decouple the temporal information into intra-gloss and inter-gloss cues and model them hierarchically via the proposed GATA module, which consists of Gloss-Aware Mask Attention~(GAMA) and Inter-Gloss Temporal Aggregation~(IGTA).
As for intra-gloss temporal aggregation, we aim to aggregate the gloss-level information from ${\bf O}^{\text{v}}$ into ${\bf O}^{\text{f}}$. To achieve this, we propose GAMA by introducing a gloss-aware mask ${\bf M} \in \mathbb{R}^{L \times P}$, 
\begin{equation}\label{GAMA}
    \text{GAMA}({\bf Q},{\bf K},{\bf V},{\bf M}) =\text{softmax}(\frac{{\bf Q}{\bf K}^{\text T}}{\sqrt{d}} \odot {\bf M}) {\bf V}
\end{equation}
where, $\bf Q$, $\bf K$, $\bf V$ are the query, key and value defined in cross-attention. $d$ is the dimension of query or key. We claim that the tokens belonging to the same class tend to have highly-relevant representations. The mask ${\bf M}$ can be considered as an attention weight, which measures the similarity between the fused and visual tokens. Thus, we first calculate the token similarity $\rho \in \mathbb{R}^{L \times P}$,
\begin{equation}\label{rho}
    \rho = \psi_{\text f}({\bf O}^{\text f}) \psi_{\text v}({\bf O}^{\text v})^{\text T}
\end{equation}
where, $\psi_{\text f}()$ and $\psi_{\text v}()$ are the linear embedding functions for the fused and visual tokens.
Also, we add a distance constrain $\delta \in \mathbb{R}^{L \times P}$ in time space to avoid computing attention between different glosses of the same category,
\begin{equation}\label{delta}
    \delta_{i,j} = {\cal K}({t}_i^{\text f}, {t}_j^{\text v}; \sigma)
\end{equation}
where ${\cal K}()$ is a Radial Basis Function~(RBF) kernel with the parameter $\sigma$. 
Since the precise timestamp for each token is not accessible, we introduce pseudo timestamps ${t}_i^{\text f}, {t}_j^{\text v} \in \mathbb{R}^1$ to represent the relative temporal position for fused and visual tokens, respectively. The pseudo timestamp for $j$-th visual token is set to $t_j^{\text v} = j$. For the fused tokens, we calculate the pseudo timestamp as the average of the pseudo timestamps of the fused tokens, defined as $t^{\text f}_i = \sum_{x=i\times\gamma}^{x=(i+1)\times\gamma} x$. Finally, the mask is defined as ${\bf M} = {\cal N}(\rho) \odot {\cal N}(\delta)$, where the $\odot$ represents the element-wise product. $\cal N$ is a zero-one normalization. 

We extend the gloss-aware mask attention into multiple heads and all the attention heads share the same mask $\bf M$. In addition, a Feed Forward Network~(FFN) module is utilized following GAMA to enhance the representation, which consists of two linear transformations with ReLU activation. Following SLT~\cite{Camgoz_CVPR20_SLT}, we add a positional encoding~(PE) process to the inputs. Thus, the intra-gloss temporal aggregation process can be summarized as follows,
\begin{equation}
    \begin{aligned}
        &{\bf Q}~= {\bf O}^{\text{f}} + {\bf P}_{\text{q}};~~{\bf K} = {\bf O}^{\text{v}} + {\bf P}_{\text{k}};~~{\bf V} = {\bf O}^{\text{v}}\\
        &{\hat{\bf O}}^{\text f} = {{\bf O}}^{\text f} + \text{GAMA}({\bf Q},{\bf K},{\bf V})\\
        &{\widetilde{\bf O}}^{\text f} = \hat{{\bf O}}^{\text f} + \text{FFN}(\hat{{\bf O}}^{\text f})\\
    \end{aligned}
\end{equation}
where ${\bf P}_{\text q} \in {\mathbb R}^{L \times C}$ and ${\bf P}_{\text k} \in {\mathbb R}^{P \times C}$ are the temporal positional encodings generated by a predefined sine function. 

After obtaining intra-gloss tokens ${\widetilde{\bf O}}^{\text f}$, we apply inter-gloss temporal aggregation to model the global motion via a multi-head self-attention~(MSA), 
\begin{equation}
\begin{aligned}
    &{\bf X}= {\widetilde{\bf O}}^{\text{f}} + {\bf P}_{\text x}\\
    &{\bf O} = {\widetilde{\bf O}}^{\text f} + \text{MSA}({\bf X})
\end{aligned}
\end{equation}
where the ${\bf P}_{\text x} \in {\mathbb R}^{L \times C}$ is also the temporal positional encoding. After applying GATA, we obtain the gloss-aware tokens ${\bf O} \in \mathbb{R}^{L \times C}$, which learns the motion cues among all glosses comprehensively and is sent to the following task heads for predicting glosses and words.

\noindent\textbf{Task heads.} Following the existing methods, we employ a classifier ${\cal F}_\text{reg}$ as Recognition Head~(RH) to predict the logits ${\bf G} = \{{\bf g}_1,{\bf g}_2,...{\bf g}_L\} \in \mathbb{R}^{L\times Y}$, where $Y$ denotes the size of the sign language vocabulary with adding a `blank' class. ${\cal F}_\text{reg}$ consists of a fully-connected layer with a softmax activation. As for handling translation task, the translation head~(TH) is to sequentially generate logits of spoken language sentences ${\bf W} = \{p(w_1), p(w_2),..., p(w_U), p(<eos>)\}$ conditioned by the gloss-aware tokens ${\bf O}$, which is an auto-regressive transformer decoder. We adopt the same translation head in ~\cite{Camgoz_CVPR20_SLT}. Due to the page limitation, Please refer to~\cite{Camgoz_CVPR20_SLT} and the supplementary material for details.

\section{Experiments}
\subsection{Datasets and evaluation protocol}

\noindent \textbf{Dataset.} We conduct analysis on synthetic~(PHOENIX14T) and real~(EvSign) event-based sign language benchmarks.
As an extension of PHOENIX14~\cite{Koller_CVIU15_PHOENIX14}, PHOENIX14T introduces German spoken language annotation and has become the primary benchmark for CSLR and SLT. It consists of 8,247 videos with a vocabulary of 1,066 and 2,887 sign and words, respectively. The number of videos for training, development and test are 7,096, 519 and 642. Based on PHOENIX14T, we build a synthetic dataset for further research using an event simulator~(V2E~\cite{Hu_CVPRW21_V2E}), where the video frames are firstly interpolated to 350 frames per second by SuperSloMo~\cite{Jiang_CVPR18_superslomo} and used for event generation. 

\noindent \textbf{Evaluation protocol.} We focus on both SLR and SLT tasks as follows:
\begin{itemize}
    \item {\bf SLR}: It only predicts the sign gloss from sign language videos without introducing any additional information. 
    Specifically, we implement a variant~($\text{Ours}_{\text{S2G}}$\footnote[2]{We use subscript to indicate which task is focused. S2G denotes the method is trained and evaluated solely on SLR task while S2GT represents the method is trained with the supervision of both gloss and spoken language and then used for SLT evaluation.}) by removing the translation branch and retrain our method for fair comparison. In this setting, we select four public-available methods including VAC~\cite{Min_ICCV21_VAC}, CorrNet~\cite{Hu_CVPR23_CorrNet}, TLP~\cite{Hu_ECCV22_TLP}, SEN~\cite{Hu_AAAI23_SEN} for comparison. All the methods are trained from scratch according to the original setting. We present WER as a metric for evaluation.
    \item {\bf SLT}: We follow the Sign2~(Gloss+Text) protocol defined in~\cite{Camgoz_CVPR20_SLT}, which is to jointly learn both recognition and translation branches in an end-to-end manner. We compare $\text{Ours}_{\text{S2GT}}$ with SLT~\cite{Camgoz_CVPR20_SLT}. Additionally, we equip the recent SLR algorithm with our translation head~(CorrNet+TH, VAC+TH), which are trained under this SLT comparison. Both ROUGE-L and BLEU-X metrics are employed to quantitatively assess these methods.
\end{itemize}
\subsection{Training Details}
The feature dimension $C$ is 1024. The downsample ratio $\gamma$ in LTF module is 4.
We adopt the widely-used CTC loss~\cite{Graves_ICML06_CTC} for SLR supervision. Inspired by VAC~\cite{Min_ICCV21_VAC}, we set an additional recognition branch ${{\cal F}'}_{\text {reg}}$ on the fused tokens ${\bf O}^{\text{f}}$. Two CTC losses ($\cal L_{\text{inter}}$ and $\cal L_{\text{final}}$) are applied to both intermediate and final outputs against the ground truth. The total loss $\cal L_{\text{SLR}}$ for SLR protocol is,
\begin{equation}
    \cal L_{\text{SLR}} = \lambda_{\text{inter}} \cal L_{\text{inter}} + \lambda_{\text{final}} \cal L_{\text{final}}
\end{equation}
Under the SLT protocol, we add a cross-entropy loss $\cal L_{\text{ce}}$ to supervise the output of translation head. Thus the overall loss of our method can be summarized as, 
\begin{equation}
    {\cal L}_\text{SLT} = \cal L_{\text{SLR}} + \lambda_{\text{ce}} \cal L_{\text{ce}}
\end{equation}
The weights for those three losses $\lambda_{\text{inter}}$, $\lambda_{\text{final}}$ and $\lambda_{\text{ce}}$ are all set to 1.
The parameters $\sigma$ in RBF is set to 16 in our experiment. The bin size of voxel grid $B$ is 5.
We adopt Adam~\cite{Ba_ICLR15_Adam} optimizer with cosine annealing strategy to adjust the learning rate.
The initial learning rate, weight decay and batch size are set to  $3e^{-5}$,  0.001 and 2, respectively. We train our method for 200 epochs to achieve convergence. 
To align the setting with existing methods, all the RGB frames are cropped to $320 \times 320$ to remove the useless background and then resized to $256 \times 256$. As for event data, we firstly generate the voxel grid. Then, the voxel grid is cropped to $480 \times 480$ and resized to $256 \times 256$.
Other competitors are trained using their own settings. Note that we modify the input channel of the first convolutional layer to fit the event input.  All the models are trained and tested on a single NVIDIA RTX 3090 GPU with 24G RAM.
\begin{table}[t]
	\renewcommand\arraystretch{1.2}   
	\caption{Comparison results for SLR on PHOENIX14T and EvSign datasets. {\bf Bold} and \underline{underline} denotes the top-two ranking performance.}
	\label{SOTA_SLR}
	\vspace{-3mm}
	\scriptsize
	\begin{center}
		\begin{tabular}{l|p{1.3cm}<{\centering}|p{1.3cm}<{\centering}p{1.3cm}<{\centering}|p{1.3cm}<{\centering}p{1.3cm}<{\centering}|p{1.5cm}<{\centering}p{1.5cm}<{\centering}}
			\bottomrule
			\multirow{2}{*}{Method} &  \multirow{2}{*}{Modal} & \multicolumn{2}{c|}{PHOENIX14T} & \multicolumn{2}{c|}{EvSign} & \multirow{2}{*}{FLOPS} & \multirow{2}{*}{Param~(M)}\\
			& & Dev~(\%) & Test~(\%) & Dev~(\%) & Test~(\%)\\
			\hline 
			VAC~\cite{Min_ICCV21_VAC} & RGB & 20.17 & 21.60 & 32.08 & 30.43 & 228.87G & 31.64 \\
			TLP~\cite{Hu_ECCV22_TLP} & RGB & 19.40 & 21.20 & 33.70 & 32.96 & 231.28G & 59.69 \\
			SEN~\cite{Hu_AAAI23_SEN} & RGB & 19.50 & 21.00 & 33.26 & 33.46 & 231.96G & 34.70 \\
			CorrNet~\cite{Hu_CVPR23_CorrNet} & RGB & 18.90 & 20.50 & 32.37 & 32.04 & 234.59G & 32.04 \\
			\hline 
			\hline 
			VAC~\cite{Min_ICCV21_VAC} & EV & 24.99 & 24.77 & 30.84 & 30.71 & \underline{238.88G} & \underline{31.65}\\
			TLP~\cite{Hu_ECCV22_TLP} & EV & 24.81 & 24.60 & 32.59 & 32.68 & 240.08G & 59.69\\
			SEN~\cite{Hu_AAAI23_SEN} & EV & 24.63 & \underline{24.51} & 33.34 & 32.71 & 242.00G & 34.70 \\
			CorrNet~\cite{Hu_CVPR23_CorrNet} & EV & \underline{24.57} & 24.55 & \underline{29.98} & \underline{29.95} & 244.63G & 32.05 \\
			\rowcolor{mygray} $\bf Ours_{\textbf{S2G}}$ & EV & \bf 23.89 & \bf 24.03 & \bf 29.19 & \bf 28.69 & \bf 0.84G & \bf 14.19\\
			\toprule
		\end{tabular}
		\vspace{-5mm}
	\end{center}
\end{table}
\subsection{Quantitative Results on Sign Language Recognition}
Table~\ref{SOTA_SLR} provides quantitative results on PHOENIX14T and EvSign datasets. Compared to existing methods that utilize RGB data, all the algorithms working with streaming event show lower WER consistently on EvSign dataset with real event, revealing the power of event stream in handling sign language recognition. However, the advantages of event data are not reflected in the results on the simulated dataset. The reason is that the video frames used for event synthesis are of poor quality with severe blur and limited frame rate.
Compared to event-based competitors, our method shows superior performance on both synthetic PHOENIX14T and EvSign datasets. Notably, our method has significant advantage in both computational cost and number of parameters, due to the concise architecture and sparse data processing.
We also evaluate the computational efficiency using FLOPS and number of parameters~(Params) on EvSign dataset\footnote[3]{Since the FLOPS and Params fluctuate based on data sparsity and sequence lengths, we calculate their averages across videos in Dev and Test sets of EvSign.}.
Compared to the most recent method~(CorrNet~\cite{Hu_CVPR23_CorrNet}), $\text{Ours}_{\text{S2G}}$ achieves 0.79\% and 1.26\% improvement with respect to WER on development and test sets of EvSign with only 0.34\% FLOPS and 44.2\% parameters.

\subsection{Quantitative Results on Sign Language Translation}
Table~\ref{SOTA_SLT_PHOENIX14T} and~\ref{SOTA_SLT_EvSign} show the comparison results for SLT on PHOENIX14T and EvSign datasets. 

As shown in Table~\ref{SOTA_SLT_EvSign}, our method achieves the best performance except BLEU-4 in the development set. On the other hand, the results on Phoenix14T~(Table~\ref{SOTA_SLT_PHOENIX14T}) cannot demonstrate the effectiveness of event camera. Compared with SLT, our method achieves 1.06\% and 0.89\% improvement in terms of ROUGE in development and test sets, respectively. Specifically, our method also exhibits significant advantages in terms of computational and parameter efficiency. Furthermore, methods with event streams are consistently better than those with RGB frames, which demonstrates the potential of event data in SLT task.

\begin{table*}[t]
	\caption{Comparison results for SLT on synthesized PHOENIX14T. (R: ROUGE-L, B-X: BLEU-X)}
	\label{SOTA_SLT_PHOENIX14T}
	\vspace{-4mm}
    \renewcommand\arraystretch{1.2}
	\scriptsize
	\begin{center}
    \resizebox{1.0\textwidth}{!}{
		\begin{tabular}{l|c|p{0.9cm}<{\centering}p{0.8cm}<{\centering}p{0.8cm}<{\centering}p{0.8cm}<{\centering}p{0.8cm}<{\centering}|p{0.9cm}<{\centering}p{0.8cm}<{\centering}p{0.8cm}<{\centering}p{0.8cm}<{\centering}p{0.8cm}<{\centering}}
			\bottomrule
            \multirow{2}{*}{Method} & \multirow{2}{*}{Modal} & \multicolumn{5}{c|}{PHOENIX-14T Dev} & \multicolumn{5}{c}{PHOENIX-14T Test} \\
		    & & R & B-1 & B-2 & B-3 & B-4 & R & B-1 & B-2 & B-3 & B-4 \\
			\hline 
		  SLT~\cite{Camgoz_CVPR20_SLT} & RGB & 41.35 & 40.06 & 29.73 & 23.24 & 18.86 & 38.69 & 38.26 & 28.36 & 22.06 & 18.06 \\
            VAC+TH~\cite{Min_ICCV21_VAC} & RGB & 39.38 & 39.22 & 29.47 & 23.44 & 19.40 & 38.57 & 39.87 & 29.53 & 23.11 & 19.00\\
            CorrNet+TH~\cite{Hu_CVPR23_CorrNet} & RGB & 39.82 & 40.05 & 30.34 & 24.04 & 19.86 & 40.26 & 41.23 & 31.42 & 24.86 & 20.48 \\
            \hline
            \hline
            SLT~\cite{Camgoz_CVPR20_SLT} & EV & \underline{39.86} & 38.99 & 29.01 & 23.54 & 18.41 & 39.21 & 39.84 & 29.25 & 23.09 & 19.23 \\
            VAC+TH~\cite{Min_ICCV21_VAC} & EV & 39.01 & 39.09 & \underline{29.64} & \bf 23.74 & \underline{19.64} & 39.34 & 39.62 & 29.55 & 23.30 & \underline{19.92}\\
            CorrNet+TH~\cite{Hu_CVPR23_CorrNet} & EV & 39.52 & \underline{39.28} & 29.60 & 23.41 & 19.24 & \underline{40.12} & \bf{40.77} & \bf{30.62} & \underline{24.11} & 19.83 \\
            \rowcolor{mygray} $\bf Ours_{S2GT}$ & EV & \bf{40.23} & \bf{39.37} & \bf{29.66} & \underline{23.67} & \bf {19.83} & \bf{40.21} & \underline{40.40} & \underline{30.47} & \bf 24.27 & \bf 20.07\\
		\toprule
		\end{tabular}
    }
	\end{center}
	\vspace{-3mm}
\end{table*}
\begin{table*}[t]
	\caption{Comparison results for SLT on EvSign. (R: ROUGE-L, B-X: BLEU-X)}
	\label{SOTA_SLT_EvSign}
	\vspace{-4mm}
    \renewcommand\arraystretch{1.2}
	\scriptsize
	\begin{center}
    \resizebox{1.0\textwidth}{!}{
		\begin{tabular}{l|c|ccccc|ccccc|cc}
			\bottomrule
            \multirow{2}{*}{Method} & \multirow{2}{*}{Modal} & \multicolumn{5}{c|}{EvSign Dev} & \multicolumn{5}{c|}{EvSign Test} & \multirow{2}{*}{FLOPS} & \multirow{2}{*}{Param~(M)} \\
		    & & R & B-1 & B-2 & B-3 & B-4 & R & B-1 & B-2 & B-3 & B-4 \\
			\hline
		  SLT~\cite{Camgoz_CVPR20_SLT} & RGB & 39.75 & 39.64 & 23.75 & 15.80 & 10.86 & 40.05 & 39.84 & 23.54 & 15.60 & 10.63 & 242.71G & 34.51 \\
            VAC+TH~\cite{Min_ICCV21_VAC} & RGB & 38.54 & 38.37 & 22.58 & 15.00 & 9.21 & 39.08 & 38.74 & 23.90 & 15.88 & 10.19 & 233.70G & 45.26 \\
            CorrNet+TH~\cite{Hu_CVPR23_CorrNet} & RGB & 38.05 & 38.74 & 22.41 & 13.98 & 9.04 & 39.41 & 39.45 & 23.74 & 15.68 & 10.57 & 239.46G & 45.66 \\
            \hline
            \hline
            SLT~\cite{Camgoz_CVPR20_SLT} & EV & \underline{39.92} & 39.06 & 23.54 & \underline{15.89} & \bf{11.21} & \underline{41.54} & 40.13 & 24.36 & 16.04 & 10.87 & 252.99G & \underline{34.51} \\
            VAC+TH~\cite{Min_ICCV21_VAC} & EV & 38.96 & 38.93 & 23.29 & 15.23 & 10.08 & 39.48 & 39.22 & 24.11 & 15.94 & 10.01 & \underline{243.84G} & 45.26 \\
            CorrNet+TH~\cite{Hu_CVPR23_CorrNet} & EV & 39.55 & \underline{39.58} & \underline{24.09} & 15.69 & 10.50 & 41.23 & \underline{40.85} & \underline{25.34} & \underline{16.95} & \underline{11.83} & 249.59G & 45.66 \\
            \rowcolor{mygray} $\bf Ours_{S2GT}$ & EV & \bf{40.98} & \bf 42.00 & \bf 25.75 & \bf 16.89 & \underline{11.20} & \bf 42.43 & \bf 41.44 & \bf 25.61 & \bf 17.55 & \bf 12.37 & \bf 6.99G & \bf 28.06 \\
		\toprule
		\end{tabular}
    } 
	\end{center}
	\vspace{-4mm}
\end{table*}

\subsection{Further Analysis}
{\noindent \bf Ablation study.}  As shown in Table~\ref{AblationAnalysis}, we conduct ablation analysis on both PHOENIX14T and EvSign. we introduce a modified VAC~\cite{Min_ICCV21_VAC} as our baseline~(B), with removing the visual alignment loss. Compared to the baseline with RGB input, the event-based baseline achieves better performance on real event, which reveals the potential of event in handling CSLR tasks. All of the proposed modules contribute positively to the recognition performance on both datasets. The sparse backbone~(SConv) can significantly drop the computational load and parameters while maintaining recognition accuracy, which can fully leverage the sparsity of event data. The simple yet effective LTF module obtains 0.47\% and 0.53\% on the test set of PHOENIX14T and EvSign, respectively. Compared to the BiLSTM, the designed GATA module can learn the temporal cues more comprehensively, leading to 0.81\% and 1.12\% WER decrease on the test sets. The final model obtains the best performance with regard to all the metrics. Compared to B(EV), the final model achieves 1.16\% and 1.39\% improvement on the test sets, which can serve as a strong baseline for further research.

\begin{figure}[t]
    \centering
    \includegraphics[width=0.95\textwidth]{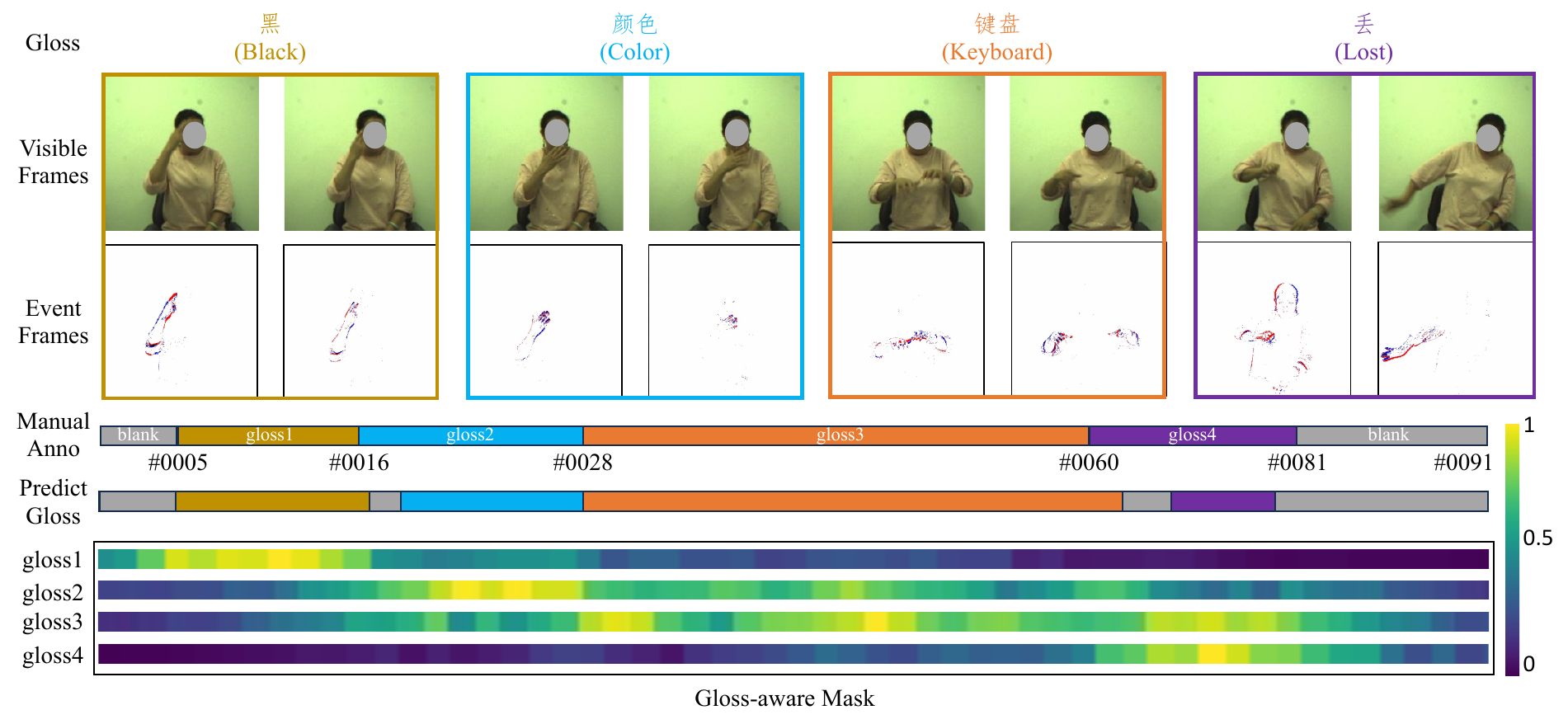}
    \caption{Visualization of the gloss-aware mask on EvSign dataset.}
    \label{fig3}
\vspace{-3mm}
\end{figure}

{\noindent \bf Visualization of gloss-aware mask.} As shown in Fig.~\ref{fig3}, we provide a visualization results of SLR task and the gloss-aware mask in GATA module.
With the guidance of GATA module, our method learns comprehensive motion cues, thus predicting all the glosses correctly. We visualize the gloss-aware mask between the tokens in corresponding gloss and visual tokens.
It demonstrates that the gloss-aware mask can precisely provide a intra-gloss correlation without any supervision, achieving gloss-aware temporal token aggregation.

\begin{table*}[t]
	\caption{Ablation analysis for SLR on PHOENIX14T and EvSign.}
	\label{AblationAnalysis}
	\vspace{-3mm}
    \renewcommand\arraystretch{1.2}
	\tiny
        \begin{center}
		\begin{tabular}{l|cc|cc|cc}
            \bottomrule
            \multirow{2}{*}{Methods} & \multicolumn{2}{c|}{PHOENIX14T} & \multicolumn{2}{c|}{EvSign} & \multirow{2}{*}{FLOPS} & \multirow{2}{*}{Params~(M)}\\
            & Dev~(\%) & Test~(\%) & Dev~(\%) & Test~(\%) \\
			\hline 
            B(RGB) & 20.38 & 21.74 & 32.00 & 30.29 & 228.87G & 31.64\\
            B(EV) & 24.92{\tiny(-4.54\%)} & 25.19{\tiny(-3.45\%)} & 30.91{\tiny(+1.09\%)} & 30.08{\tiny(+0.27\%)} & 238.88G & 31.65\\
            B(EV)+SConv & 24.68{\tiny(+0.24\%)} & 25.38{\tiny(-0.19\%)} & 30.21{\tiny(+0.70\%)} & 30.01{\tiny(+0.07\%)} & 1.22G & 21.90\\
            B(EV)+SConv+LTF & 24.33{\tiny(+0.35\%)} & 24.91{\tiny(+0.47\%)} & 29.79{\tiny(+0.42\%)} & 29.48{\tiny(+0.53\%)}  & 0.72G & 13.13 \\
            B(EV)+SConv+GATA & 24.04{\tiny(+0.64\%)} & 24.57{\tiny(+0.81\%)} & 29.41{\tiny(+0.80\%)} & 28.89{\tiny(+1.12\%)} & 1.69G & 26.10\\
            B(EV)+SConv+LTF+GATA & 23.89{\tiny(+0.15\%)} & 24.03{\tiny(+0.54\%)} & 29.19{\tiny(+0.22\%)} & 28.69{\tiny(+0.20\%)} & 0.84G & 14.19\\
    		\toprule
		\end{tabular}
	\end{center}
	\vspace{-4mm}
\end{table*}

\begin{table}[t]
    \centering
    \caption{Ablation analysis of the proposed modules on EvSign dataset.} \label{module_analysis}
    \vspace{-3mm}
    \renewcommand\arraystretch{1.2}
    \begin{subtable}{.4\textwidth}
        \centering
        \caption{\scriptsize Ablation analysis of LTF.}
        \resizebox{0.92\textwidth}{!}{
        	\vspace{-1mm}
        \begin{tabular}{l|cc}
			\bottomrule
            \multirow{2}{*}{} & \multicolumn{2}{c}{EvSign(EV)}\\
             & Dev~(\%) & Test~(\%) \\
			\hline 
            w/o aggregation & 40.95 & 40.34\\
            MaxPooling & 34.44 & 33.24\\
            AvgPooling & 32.27 & 31.56\\
            1D-CNN~\cite{Min_ICCV21_VAC} & \underline{29.41} & \underline{28.89}\\
            LTF(Ours) & \bf 29.19 & \bf 28.69\\ 
            \toprule
		\end{tabular}
        }
    \end{subtable} 
    \begin{subtable}{.4\textwidth}
        \centering
        \caption{\scriptsize Ablation analysis of GATA.}
        \resizebox{1.0\textwidth}{!}{
        	\vspace{-1mm}
        \begin{tabular}{l|cc}
                \bottomrule
                \multirow{2}{*}{} & \multicolumn{2}{c}{EvSign(EV)}\\
                 & Dev~(\%) & Test~(\%) \\
                \hline
                w/o GAMA & 30.38 & 31.14\\
                $\delta$-only & 29.89 & 29.83\\
                $\rho$-only & \underline{29.63} & 29.74\\
                GAMA-hard & 29.74 & \underline{29.31}\\
                GAMA-Soft~(Ours) & \bf 29.19 & \bf 28.69\\
                \toprule
            \end{tabular}
        }
    \end{subtable}
   \vspace{-3mm}
\end{table}

{\noindent \bf Analysis of token aggregation strategy.} 
We compare several token aggregation strategies and the performance on EvSign dataset is shown in Table~\ref{module_analysis}(a). We implement the `w/o aggregation' by removing the selection module, where the visual tokens are directly sent to GATA modules to output the probability of gloss. The inferior performance indicates the necessity of token aggregation module. The useless information may lead to a delete or insert error, which significantly affects the recognition performance. We also compare the simplest selection strategies, which are denoted as MaxPooling and AvgPooling. Those methods can decrease the WER, while selection in a soft manner~(AvgPooling) is better than the hard one~(MaxPooling). This indicates the local fusion is necessary for learning discriminative tokens for temporal modeling. Both 1D-CNN designed in~\cite{Min_ICCV21_VAC} and our method apply local aggregation before selection, leading to favorable performances. Our method achieves 0.22\% and 0.2\% improvement than 1D-CNN, which shows the effectiveness of LTF module.

{\noindent \bf Analysis of temporal aggregation module.} Temporal information module is the key component in sign language tasks. To this end, we compare various methods to demonstrate the capabilities of our method in temporal modeling. As shown in Table~\ref{module_analysis}(b), we set our method without intra-gloss temporal aggregation~(GAMA) as baseline. The method solely relies on inter-gloss temporal modeling via global self-attention. $\rho$-only and $\delta$-only GAMA denote that the mask $\bf M$ in Eq.~(\ref{GAMA}) is set to $\rho$ and $\delta$ in Eq.~(\ref{rho}) and Eq.~(\ref{delta}). Compared with the baseline, $\rho$-only and $\delta$-only GAMA achieve 1.4\% and 1.31\% performance gain in the test set. Furthermore, we compare the effectiveness of soft and hard mask in GAMA. GAMA-hard means that we binarize the learned $\bf M$ with a threshold $1e^{-3}$. we find that the hard mask will lead to a performance decrease. Soft manners exhibit greater flexibility, resulting in improved aggregation performance.

\section{Conclusion}
In this paper, we unveil the power of events in sign language tasks. We first build a comprehensive dataset for event-based sign language recognition and translation. The dataset contains more than 6.7K sign videos captured by high-quality event camera, which covers most of daily topics. It can greatly promote the development of event-based vision and sign language related tasks. Furthermore, we propose a transformer-based framework for both tasks by fully exploiting the sparsity and high temporal resolution characteristics of events. The proposed gloss-aware temporal aggregation module could effectively model temporal information in a global-and-local manner and a gloss-aware representation is computed for SLR and SLT tasks. Our method shows favorable performance on the sign language datasets with synthetic and real events with only 0.34\% FLOPS and 44.2\% network parameters.
~\\
~\\
{ \normalsize \bf \noindent Acknowledgment}\\
The research was partially supported by the National Natural Science Foundation of China, (grants No. 62106036, U23B2010, 62206040, 62293540, 62293542), the Dalian Science and Technology Innovation Fund (grant No. 2023JJ11CG001).

\clearpage
\bibliographystyle{splncs04}
\bibliography{main}
\end{document}


\begin{center}
{\bf {\Large -- Supplementary Material -- \\
{\color{red}E}{\color{blue}v}Sign: Sign Language Recognition and Translation with Streaming Events}}
\end{center}

\section{Translation Head}
Following~\cite{Camgoz_CVPR20_SLT}, we introduce an auto-regressive transformer decoder as our translation head, which contains four decoder blocks. Each decoder block consists of a masked self-attention layer, an encoder-decoder attention layer and a feed-forward layer. The spoken language sentence is first prefixed with a special beginning-of-sentence token~(<bos>). Then, the generated word tokens are sent to the masked self-attention layer, where each token can only use its predecessors to extract contextual representation. The encoder-decoder attention layer is used to learn the mapping between gloss-aware and word tokens. Finally, a feed-forward layer is appended to predict the probability of words in spoken language. The translation head learns to generate target sentence in an auto-regressive manner until it produces a special end-of-sentences~(<eos>).

\section{Comparison between Synthetic and Real Data}

\begin{figure}[t]
    \centering
    \includegraphics[width=0.9\textwidth]{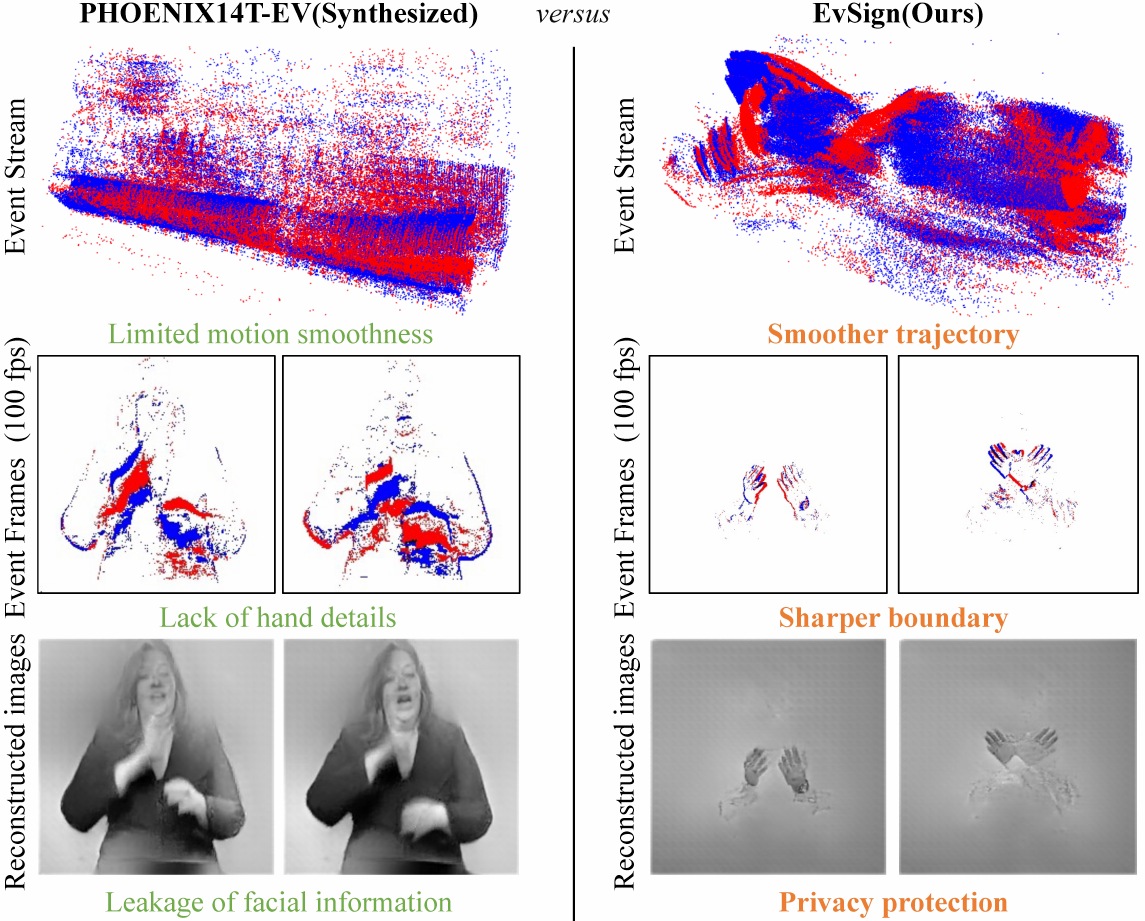}
    \caption{Comparison between synthetic and real Data.}
    \label{fig1_supp}
\end{figure}

As shown in Fig.~\ref{fig1_supp}, we claim that there is a gap between synthetic and real events. We conclude the advantages of the collected real event dataset EvSign from three aspects.

First, the synthetic events from RGB frames suffer from poor continuity, which relates to the framerate of RGB videos. In contrast, our dataset, captured by high-quality event cameras, can generate rich events within mircosecond response, thus providing smoother trajectories.
Second, as illustrated in the second row, the RGB frames are blurry in fast motion scenarios. The blurriness leads to the loss of informative boundaries in event data, which are crucial cues for sign language tasks. However, the event frames sampled in real events contain sharper boundaries, which can provide more discriminative details, achieving better recognition results. As shown in Sec. 5 of the main submission, methods utilizing real events yield superior performance compared to those relying on RGB inputs. This shows the effectiveness of real events in sign language recognition and translation tasks.
Third, we utilize an event-based image reconstruction method~(E2VID\cite{Rebecq_PAMI21_E2VID}) to reconstruct intensity images from events, shown in the third row of Fig.~\ref{fig1_supp}. Given little movement with signer's head, the reconstruction method cannot recover signer's many facial details, thereby protecting the user's privacy.
 
 \section{Comparison on SL-Animals-DVS} 
 Since EvASL is not publicly available, we conduct comparison on SL-Animals-DVS for ISLR. Please note that we focus on CSLR, which is different from ISLR. Thus, we modify those methods by replacing CTC loss with cross-entropy loss and aggregating features along temporal dimension to perform sequence-level prediction rather than frame-level prediction. Following previous methods, we adopt 4-fold cross-validation and report the mean and standard deviation of accuracy~(Acc.) in Tab.~\ref{SL-Animals-DVS}. Our method outperforms other methods with 3.45\% improvement.

 \begin{table}[h]
 \caption{\footnotesize Results on SL-Animals-DVS dataset.}
 \label{SL-Animals-DVS}

 \begin{center}
 	\resizebox{0.95\textwidth}{!}{
 		\begin{tabular}{ccccccc}
 			\toprule
 			&  VAC & TLP & SEN & CorrNet & Ours\\
 			\midrule
 			Acc.(\%) & 95.48$\pm$1.15 & 96.23$\pm$0.46 & 96.05$\pm$0.82 & 95.38$\pm$1.10 & \bf 99.68$\pm$\bf0.09\\    
 			\bottomrule
 		\end{tabular}
 	}
 \end{center}
 \end{table}
 
\section{More Results on Sign Language Recognition}
We also conduct a comparison of sign language recognition~(SLR) task on the synthetic CSL-Daily dataset, shown in Table~\ref{SOTA_SLR_supp}. We note that the synthetic events do not bring gains to SLR methods due to the limited smoothness and blurry content with the original frame sequences. This also reflects the necessity for proposing a high-quality sign language dataset with real events. Compared to the state-of-the-art method~(CorrNet~\cite{Hu_CVPR23_CorrNet}), our method achieves 0.81\%/0.88\% WER reduction in both Dev and Test subsets, which demonstrates its effectiveness in handling sign language recognition.

\begin{table}[h]
    \renewcommand\arraystretch{1.2}   
	\caption{Comparison results for SLR on CSL-Daily dataset.}
    \vspace{-5mm}
    \label{SOTA_SLR_supp}
	\scriptsize
	\begin{center}
		\begin{tabular}{c|cccc|ccccc}
			\bottomrule
            Modal &  \multicolumn{4}{c|}{RGB} &  \multicolumn{5}{c}{EV} \\
            \hline
            Method & VAC~\cite{Min_ICCV21_VAC} & TLP~\cite{Hu_ECCV22_TLP} & SEN~\cite{Hu_AAAI23_SEN} & CorrNet~\cite{Hu_CVPR23_CorrNet} & VAC~\cite{Min_ICCV21_VAC} & TLP~\cite{Hu_ECCV22_TLP} & SEN~\cite{Hu_AAAI23_SEN} & CorrNet~\cite{Hu_CVPR23_CorrNet} & $\text{Ours}_{\text{S2G}}$\\
            \hline
            Dev(\%) & 31.24 & 32.30 & 31.10 & 30.60 & 35.16 & 36.10 & 35.42 & \underline{34.81} & \bf 34.00\\ 
            Test(\%) & 30.68 & 32.35 & 30.70 & 30.10 & 34.75 & 36.18 & 35.27 & \underline{34.70} & \bf 33.82\\ 
            \toprule
		\end{tabular}
	\end{center}
 \vspace{-8mm}
\end{table}
\section{More Visualization of Gloss-Aware Mask}
\begin{figure}[t]
    \centering
    \includegraphics[width=1.0\textwidth]{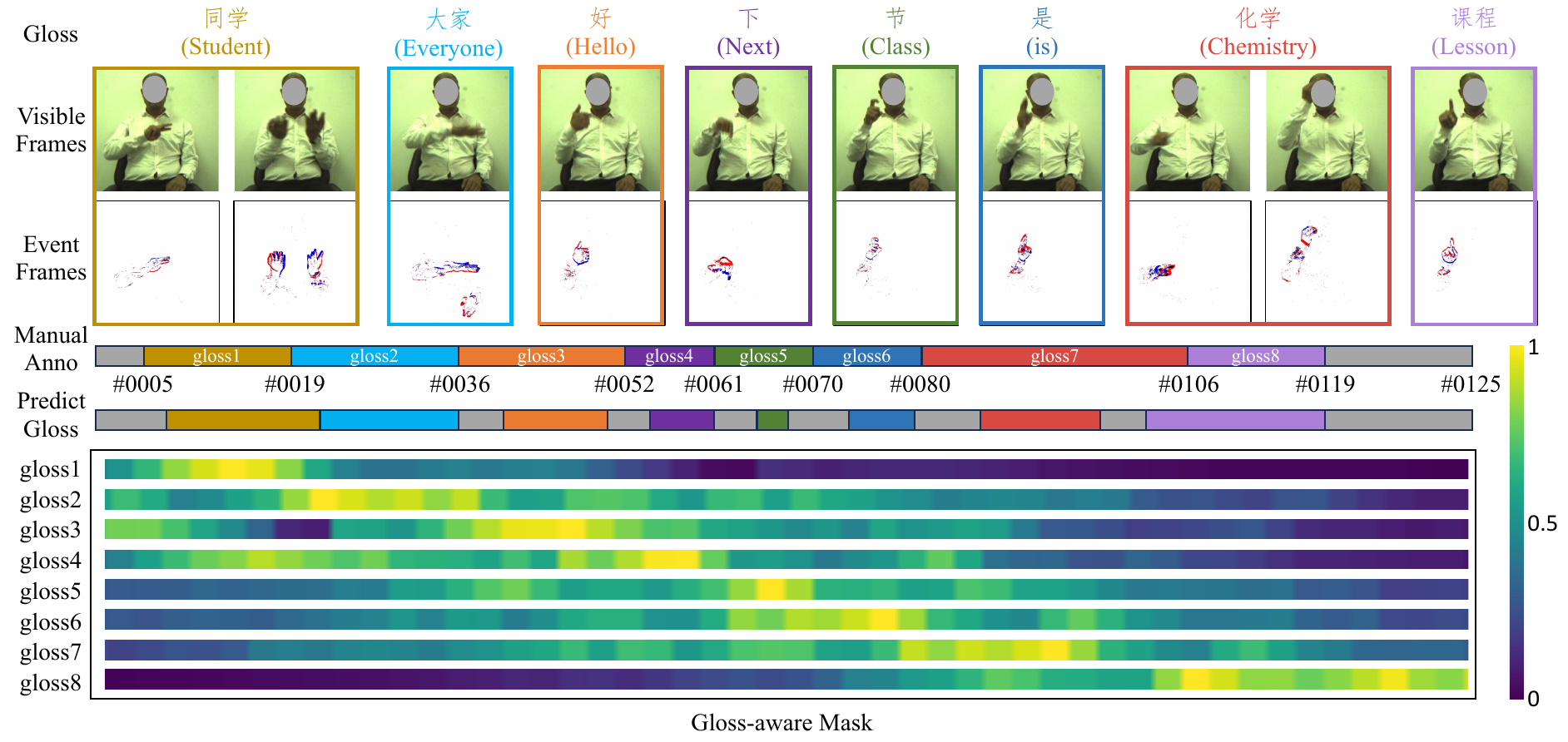}
    \caption{Comparison between synthetic and real Data.}
    \label{fig2_supp}
\end{figure}

In this section, we aim to conduct qualitative analysis on the proposed Gloss-Aware Mask Attention~(GAMA). Fig.~\ref{fig2_supp} illustrates the learned mask in GAMA.
We claim that it is capable of adaptively aggregating tokens that belong to the same gloss, thus modeling the complete sign language gestures and eliminating the interference from different glosses.

\section{Analysis on weight of loss components} 
Following VAC and CorrNet, we set all the weights to 1 without modification. We conduct analysis on EvSign to evaluate the influence of loss weights. As shown in Tab.~\ref{loss_weight}, our method is not sensitive to variations in loss weights.

\begin{table}[h]
\caption{\footnotesize Analysis of loss weights on EvSign dataset.}
\label{loss_weight}

\begin{center}
	\resizebox{0.5\textwidth}{!}{
		\begin{tabular}{cccccccccc}
			\toprule
			$\lambda_{\text{inter}}$ & 1 & 1 & 1 & 5 & 5 & 10 & 10\\
			$\lambda_{\text{final}}$ & 1 & 5 & 10 & 1 & 10 & 1 & 5\\
			\midrule
			Dev~(\%) & \bf 29.19 & 29.43 & 29.27 & 29.33 & 29.24 & 29.59 & 30.17\\
			Test~(\%) & 28.69 & 29.37 & 29.01 & 29.07 & \bf 28.58 & 28.96 & 30.22\\
			\bottomrule
		\end{tabular}
	}
\end{center}
\end{table}

\bibliographystyle{splncs04}
\bibliography{main}